\pdfoutput=1

\documentclass[11pt]{article}

\usepackage[]{ACL2023}

\usepackage{times}
\usepackage{latexsym}

\usepackage[T1]{fontenc}
\usepackage[utf8]{inputenc}

\usepackage{microtype}
\usepackage{inconsolata}

\usepackage{graphicx}

\usepackage{multirow,multicol}
\usepackage{booktabs}
\usepackage{siunitx}
\usepackage{makecell}

\usepackage{amssymb}
\usepackage{amsmath}  
\usepackage{bm}
\usepackage[linesnumbered,ruled,vlined]{algorithm2e}

\newcommand{\ignore}[1]{}
\newcommand{\fref}[1]{Fig.~\ref{#1}}
\newcommand{\tref}[1]{Table \ref{#1}}
\newcommand{\sref}[1]{\S\ref{#1}}
\newcommand{\aref}[1]{\ref{#1}}

\title{Analyzing Text Representations by Measuring Task Alignment}

\author{
  {\bf Cesar Gonzalez-Gutierrez},
  {\bf Audi Primadhanty},
  {\bf Francesco Cazzaro},
  {\bf Ariadna Quattoni} \\
  Universitat Politècnica de Catalunya, Barcelona, Spain \\
  \texttt{\{cesar.gonzalez.gutierrez,
  audi.primadhanty, francesco.cazzaro\}@upc.edu},\\
  \texttt{aquattoni@cs.upc.edu}
}

\begin{document}

\maketitle

\begin{abstract}
Textual representations based on pre-trained language models are key, especially in few-shot learning scenarios. What makes a representation good for text classification? Is it due to the geometric properties of the space or because it is well aligned with the task? We hypothesize the second claim. To test it, we develop a task alignment score based on hierarchical clustering that measures alignment at different levels of granularity.
Our experiments on text classification validate our hypothesis by showing that task alignment can explain the classification performance of a given representation.
\end{abstract}

\section{Introduction}

Recent advances in text classification have shown that representations based on pre-trained language models are key, especially in few-shot learning scenarios
\cite{ein-dor-etal-2020-active, lu_investigating_2019}. It is natural to ask: What makes a representation good for text classification in this setting? Is the representation good due to intrinsic geometric properties of the space or because it is well \emph{aligned} with the classification task? The goal of this paper is to answer this question to better understand the reason behind the performance gains obtained with pre-trained representations.

Our hypothesis is that representations better aligned with class labels will yield improved performance in few-shot learning scenarios. The intuition is simple: in this setting, the limited number of labeled samples will only provide a sparse coverage of the input domain. However, if the representation space is properly aligned with the class structure, even a small sample can be representative. To illustrate this, take any classification task. Suppose we perform clustering on a given representation space that results in a few pure clusters (with all samples belonging to the same class). Then, any training set that `hits' all the clusters can be representative. Notice that there is a trade-off between the number of clusters and their purity. A well-aligned representation is one for which we can obtain a clustering with a small number of highly pure clusters. Based on this, we propose a task alignment score based on hierarchical clustering that measures alignment at different levels of granularity: Task Hierarchical Alignment Score (\textsc{Thas}).

\begin{figure}[t]
    \centering
    \includegraphics[width=\linewidth]{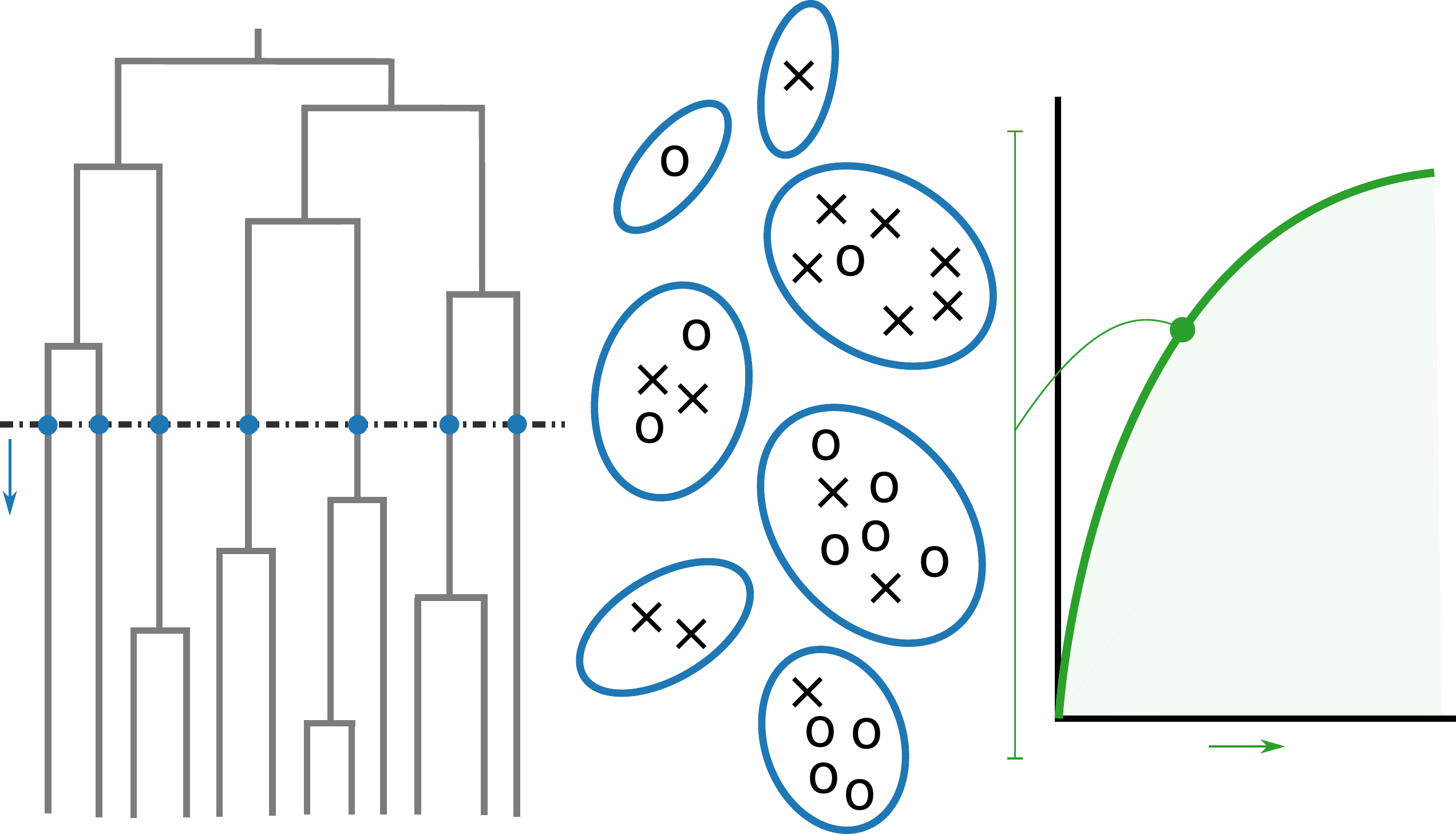}
    \caption{Three-step process for computing \textsc{Thas}.}
    \label{fig:concept}
\end{figure}

To test our hypothesis that task alignment is key we conduct experiments on several text classification datasets comparing different representations. Our results show that there is a clear correlation between the \textsc{Thas} of a representation and its classification performance under the few-shot learning scenario, validating our hypothesis and showing that task alignment can explain performance. 
In contrast, our empirical study shows that intrinsic geometric properties measured by classical clustering quality metrics fail to explain representation performance in the few-shot learning scenario.

Our study suggests an answer to our main question: A good efficient representation (i.e. one that enables few-shot learning) is a representation that induces a good alignment between latent input structure and class structure. Our main contributions are: 1) We develop a score based on hierarchical clustering (\sref{thas}) that measures the extent to which a representation space is aligned with a given class structure and 2) We conduct an empirical study using several textual classification datasets (\sref{experiments}) that validates the hypothesis that the best representations are those with a latent input structure that is well aligned with the class structure.

\section{Task Hierarchical Alignment Score}
\label{thas}

We now present the Task Hierarchical Alignment Score (\textsc{Thas}) designed to measure the alignment between a textual representation and the class label for a given task. The idea is quite simple, in a good representation space, points that are close to each other should have a higher probability of belonging to the same class. Therefore, we could perform clustering of the points and obtain \emph{high purity} clusters, where most points belong to the same class.
We assume that we are given: a dataset $S=\{(\bm{x}_i, y_i)\}_{i=1}^n$ of $n$ labeled data points where $\bm{x} \in \mathcal{X}$ is a text fragment and $y \in \mathcal{Y}$ its corresponding class label (e.g., a sentiment classification label) and a representation function $r: \mathcal{X} \to \mathbb{R}^d$ mapping points in $\mathcal{X}$ to a  $d$-dimensional representation space $\mathbb{R}^d$ (e.g., a sparse bag-of-words).

Our goal is to compute a metric $\tau(S, r)$ that takes some labeled domain data and a representation function and computes a real value score.
\fref{fig:concept} illustrates the steps involved in computing \textsc{Thas}. There are three main steps: 1) hierarchical clustering, 2) computing clustering partition alignments, and 3) computing the aggregate metric.
In the first step, we compute the representation of each point and build a data dendrogram using hierarchical clustering. The data dendrogram is built by merging clusters, progressively unfolding the latent structure of the input space. Traversing the tree, for each level we get a partition of the training points into $k$ clusters. In step 2, for each partition, we measure its alignment with the class label distribution producing an alignment curve as a function of $k$. Finally, we report the area under this curve.
Algorithm \ref{alg:computing_thas} summarizes the whole procedure.
Implementation details and performance information can be found in \ref{thas_implementation}.

\subsection{Hierarchical Clustering}

In the first step, we will consider the input points
$\bm{X} = \{\bm{x}_i \mid (\bm{x}_i, y_i) \in S \}$
and the representation function $r$ to obtain a representation of all points
$\bm{R} = \{r(\bm{x}_i) \mid \bm{x}_i \in \bm{X}\}$.

We then apply Hierarchical Clustering (HC) to the  points in $\bm{R}$ obtaining a dendrogram $\mathcal{D} = \text{HC}(\bm{R}) = \{\mathcal{P}_k\}_{k=1}^n$ that defines a set of $n$ cluster partitions. \fref{fig:concept} (left) shows a diagram of a dendrogram. The root of this tree is the whole set and, at the leaves, each point corresponds to a singleton. At intermediate levels, top-down branching represents set splitting.

For each level $k = 1, \ldots, n$ of the dendrogram there is an associated clustering partition of the input points into $k$ clusters $\mathcal{P}_k = \{C_j\}_{j=1}^k$. That is, for any particular level we have a family of $k$ non-empty disjoint clusters that cover the representation $\bm{R} = \bigcup_{j=1}^k C_j$, where each representation point $r(\bm{x}) \in \bm{R}$ is assigned to one of the $k$ clusters.

\begin{algorithm}[t]
\caption{\textsc{Thas}}
\label{alg:computing_thas}\DontPrintSemicolon
    \SetKw{In}{in}
    \SetKw{Where}{where}
    \KwIn{
        Dataset $S=\{(\bm{x}_i, y_i)\}_{i=1}^n$,
        representation function $r$
    }
    \KwOut{$\tau(S,r)$}
    Get representation: \newline
    $\bm{R} = \{r(\bm{x}_i) \mid \bm{x}_i \in \bm{X}\}$ \\
    Run Hierarchical Clustering:
    $\mathcal{D} = \text{HC}(\bm{R}) = \{\mathcal{P}_k\}_{k=1}^n$ \\
    Traverse the dendrogram: \newline
    \ForEach{partition $\mathcal{P}_k \subset \mathcal{D}$}{
        Predict scores for all points: \newline
        \ForEach{point $\bm{x}_i \in \bm{X}$ \In $i = 1, \ldots, n$ \Where $r(\bm{x}_i) \in C \subset \mathcal{P}_k$}{
            Label prediction scores: \newline
            \lForEach{$y'_j \in \mathcal{Y}$ \In $j = 1,\ldots, |\mathcal{Y}|$}{
                $\hat{\bm{Y}}_{k, i, j} = s(\bm{x}_i, y'_j)$ 
            }
        }
        Partition alignment score: \newline
        $a(\mathcal{P}_k) = \text{AUC}_{y^+}(\hat{\bm{Y}}_k, \bm{Y})$
    }
    Final aggregate metric: \newline
    $\tau(S, r) = \frac{1}{n} \sum_{k=1}^n a(\mathcal{P}_k)$
\end{algorithm}

\begin{table*}[t]
\setlength{\tabcolsep}{5.5pt}
\centering
\begin{tabular}{l ccccc ccccc ccccc}
\toprule
\multirow{2}{*}{\textbf{Repr.}} & \multicolumn{5}{c}{\textbf{ALC}} & \multicolumn{5}{c}{\textbf{\textsc{Thas}}} & \multicolumn{5}{c}{\textbf{ADBI}} \\
\cmidrule(lr){2-6} \cmidrule(lr){7-11} \cmidrule(lr){12-16}
& IM & WT & CC & S1 & $\mu$ & IM & WT & CC & S1 & $\mu$ & IM & WT & CC & S1 & $\mu$ \\
\midrule
BERT\textsubscript{all} & \emph{.84} & \emph{.50} & \emph{.32} & \emph{.79} & \bf{.61} & \emph{.84} & \emph{.67} & \emph{.27} & \emph{.75} & \bf{.63} & 2.87 & 3.03 & 3.31 & 3.25 & 3.11 \\
GloVe & .80 & .48 & .26 & .74 & .57 & .80 & .63 & .26 & .73 & .60 & \emph{2.62} & \emph{2.12} & 2.01 & \emph{2.47} & \textbf{2.31} \\
BERT\textsubscript{cls} & .80 & .48 & .23 & .74 & .56 & .80 & .56 & .22 & .74 & .58 & 2.81 & 2.97 & 3.15 & 2.92 & 2.96 \\
fastText & .75 & .41 & .18 & .66 & .50 & .77 & .57 & .21 & .71 & .56 & 2.78 & 2.13 & \emph{1.93} & \emph{2.47} & 2.33 \\
BoW   & .76 & .32 & .11 & .59 & .45 & .71 & .50 & .20 & .68 & .52 & 3.14 & 3.83 & 4.23 & 3.86 & 3.76 \\
\bottomrule
\end{tabular}
\caption{Learning curve performance (ALC), task alignment (\textsc{Thas}), and unsupervised clustering quality (ADBI) for different representations and datasets. (Rows are sorted by average ALC.)}
\label{tab:results}
\end{table*}

\subsection{Partition Alignment Score}

We use the gold labels $\bm{Y} = \{y_i \mid (\bm{x}_i, y_i) \in S\}$ to compute an alignment score $a(\mathcal{P}_k)$ for each partition $\mathcal{P}_k \subset \mathcal{D}$. We compute it in two parts.

First, for every point $\bm{x} \in \bm{X}$ and label $y' \in \mathcal{Y}$ we compute a label probability score by looking at the gold label distribution of the cluster $C$ to which the point belongs in the clustering partition:
\begin{equation}
s(\bm{x}, y') = \frac{1}{|C|} \#[y' \in C]
\end{equation}

where $\#[y' \in C]$ is the number of samples in cluster $C$ with gold label $y'$. Intuitively, this assigns to a point $\bm{x}$ a label probability that is proportional to the distribution of that label in the cluster $C$.

Second, we use the label probability scores of all points $\hat{\bm{Y}}_k = \{s(\bm{x}_i, y'_j) \mid \bm{x}_i \in \bm{X}, y'_j \in \mathcal{Y}\}$ and the dataset gold labels $\bm{Y}$ to compute a partition alignment score.
We choose as a single metric the area under the precision-recall curve (AUC) because it has the nice property that it applies to tasks with both balanced and unbalanced class distributions.\footnote{F1 could be a valid alternative, but this metric requires the validation of decision thresholds.}
More specifically, we compute the AUC of the target (positive) class $y^+ \in \mathcal{Y}$ of the dataset (more details in the experimental part in \sref{experiments}):
\begin{equation}
a(\mathcal{P}_k) = \text{AUC}_{y^+}(\hat{\bm{Y}}_k, \bm{Y})
\end{equation}

\subsection{Final Aggregate Metric: \textsc{Thas}}

Once we have an alignment score for every level of the hierarchical dendrogram, we are ready to define our final Task Hierarchical Alignment Score (\textsc{Thas}). Consider the alignment scoring function $a$ applied to the partition corresponding to the lowest level of the dendrogram. The alignment score will be $a(\mathcal{P}_n) = 1$ because every cluster in this partition is a singleton and therefore $\#[y' \in C]$ will be $1$ for the gold label and $0$ for any other label.
At the other end, for the partition corresponding to the root of the dendrogram (where all points belong to a single cluster), the alignment score $a(\mathcal{P}_1)$ is the AUC corresponding to assigning to every point $\bm{x} \in \bm{X}$ a prediction score for each label $y' \in \mathcal{Y}$ equal to the relative frequency of $y'$ in $\boldsymbol{Y}$.

Consider now the alignment score as a function of the size of the partition. As we increase $k$ we will get higher scores. A good representation is one that can get a high score while using as few clusters as possible. 
Instead of choosing a predefined level of granularity, we propose to leverage the alignment information across all levels.
To achieve this, we consider the alignment score as a function of the number of clusters and measure the area under $a(\mathcal{P}_k)$.\footnote{
We could consider weighting methods that neutralize uninformative areas in the curve. In particular, we could subtract the scores originating from a random clustering. However, this contribution is solely determined by the sample size and the prior distribution. As a result, it would not have any impact when comparing representations.}
We are ready to define our final metric:
\begin{equation}
\tau(S, r) = \frac{1}{n} \sum_{k=1}^n a(\mathcal{P}_k)
\end{equation}

\section{Experimental Setup}
\label{experiments}

In this section we empirically study the correlation of few-shot learning performance with 1) \textsc{Thas} and 2) an unsupervised clustering quality metric.

We use four text classification datasets with both balanced and imbalanced label distributions: IMDB \citep[IM;][]{maas-etal-2011-learning}, WikiToxic \citep[WT;][]{wulczyn_ex_2017}, Sentiment140 \citep[S1;][]{maas-etal-2011-learning} and CivilComments \citep[CC;][]{borkan_nuanced_2019}.

We will compare the following representations: a sparse bags-of-words (BoW); BERT embeddings \citep{devlin-etal-2019-bert} using two token average pooling strategies (BERT\textsubscript{all} and BERT\textsubscript{cls}); GloVe \citep{pennington_glove_2014}; and fastText \citep{bojanowski_enriching_2017, joulin_bag_2016}.

For further details, please refer to \aref{experimental_details}.

\subsection{Few-Shot Performance vs. \textsc{Thas}} 

Since the focus of these experiments is comparing representations, we follow previous work on probing representations and use a simple model \cite{tenney_what_2019, lu_investigating_2019}. More precisely, we use a linear max-entropy classifier trained with $l2$ regularization.
\ignore{
Complementary experiments presented in \ref{representation_is_key} show that, in the few-shot learning scenario, linear models can be competitive if they are provided with a good representation, performing similarly to more complex deep models.
}

To simulate a few-shot learning scenario, we create small training sets by selecting $N$ random samples, from $100$ to $1000$ in increments of $100$. For each point $N$ in the learning curve we create an 80\%/20\% 5-fold cross-validation split to find the optimal hyper-parameters. We then train a model using the full $N$ training samples and measure its performance on the test set. We repeat the experiment with 5 random seeds and report the mean results. As the evaluation metric, we use accuracy for the balanced datasets (IMDB and Sentiment140) and F1 for the imbalanced datasets (WikiToxic and CivilComments).

We generate learning curves for each dataset and representation (\ref{curves}). To study the correlation between task alignment and few-shot learning performance, it is useful to have a single score that summarizes the learning curve: We use the area under the learning curve (ALC). Representations with a larger ALC perform better in the few-shot learning scenario.\footnote{Alternatively, we could have picked a single point but we believe that ALC provides a more robust measure of few-shot learning performance and allows for a more concise analysis.}
We observe that BERT\textsubscript{all} is consistently the best representation followed by BERT\textsubscript{cls} and GloVe performing similarly. Representations based on word embeddings are better than the sparse baseline for all datasets, except for fastText which does not exhibit a consistent improvement.

To test for correlation, we also computed \textsc{Thas} for each representation and dataset. (The corresponding curves can be found in \ref{curves}.)
Since this metric is a measure of the alignment between a label distribution and an input representation, there is a \textsc{Thas} score per label.\footnote{We could also aggregate the scores of different classes, for example taking the average of the scores over all labels.}
In the classification tasks that we consider there is always a single target class (e.g., toxicity for WikiToxic). We measure the alignment score with respect to this class.

\begin{figure}[t]
    \centering
    \includegraphics[width=0.85\linewidth]{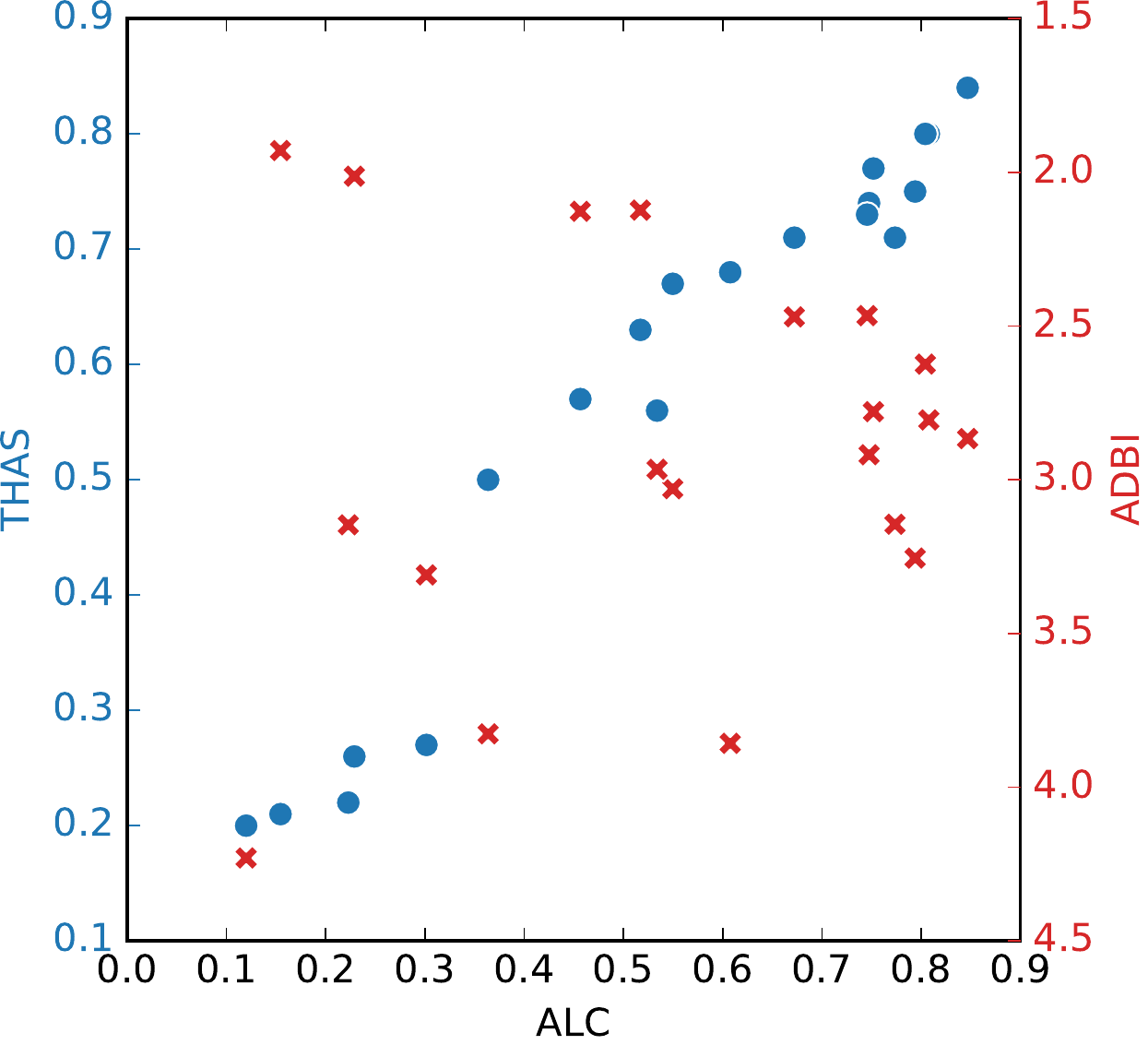}
    \caption{Few-shot performance (ALC) vs. task alignment (\textsc{Thas}) and clustering quality (ADBI).
    }
    \label{fig:correlations}
\end{figure}

\tref{tab:results} summarizes the results showing ALC (left) and corresponding \textsc{Thas} (center) for all representations and datasets. Overall, BERT\textsubscript{all} is the best representation for few-shot learning followed by GloVe and BERT\textsubscript{cls}. All the representations based on pre-trained word embeddings significantly outperform the baseline sparse BoW representation.
\textsc{Thas} predicts accurately the relative ranking between representations and the larger gap between BERT\textsubscript{all} and the rest. \fref{fig:correlations} shows a scatter plot of \textsc{Thas} as a function of ALC (blue dots; each point corresponds to a dataset and representation). We compute the correlation coefficients, which are displayed in \tref{tab:correlation}. We observe a clear positive correlation between the two metrics, providing supporting evidence for our main hypothesis that a good representation under few-shot learning is a representation that is well aligned with the classification task.

\begin{table}[t]
    \centering
    \begin{tabular}{l l l}
        \toprule
        ($\mu$)ALC vs & $r_p$ (p-value) & $r_s$ (p-value) \\
        \midrule
        \textsc{Thas} & $0.98$ ($<10^{-12}$) & $0.99$ ($<10^{-17}$) \\
        ADBI & $0.11$ ($0.62$) & $0.07$ ($0.76$) \\
        $\mu$\textsc{Thas} & $0.98$ ($0.002$) & $1.0$ ($0.017$) \\
        $\mu$ADBI & $-0.41$ ($0.48$) & $-0.3$ ($0.68$) \\
        \bottomrule
    \end{tabular}
    \caption{Pearson correlation coefficient ($r_p$) and  Spearman's correlation coefficient ($r_s$) with the corresponding p-values for ALC vs.\,\textsc{Thas} and ALC vs.\,ADBI, and similar analysis for mean scores across all datasets.}
    \label{tab:correlation}
\end{table}

\subsection{Unsupervised Clustering Quality}
\label{dbi}

We now look at standard metrics of cluster quality and test if they can explain few-shot learning performance. We use the \citet{davies_cluster_1979} index (DBI) to measure the quality of the cluster partitions at every level of the dendrogram. This metric measures the compactness of each cluster and their separation, with better cluster partitions scoring lower. Similar to the computation of \textsc{Thas} described in \sref{thas}, we compute DBI as a function of the number of clusters $k$ corresponding to each level of the dendrogram.
As an aggregate metric, we calculate the area under these curves to obtain a single ADBI score.
(The curves are shown in \ref{curves}.)

\ignore{
\begin{figure}[t]
    \centering
    \includegraphics[width=0.85\linewidth]{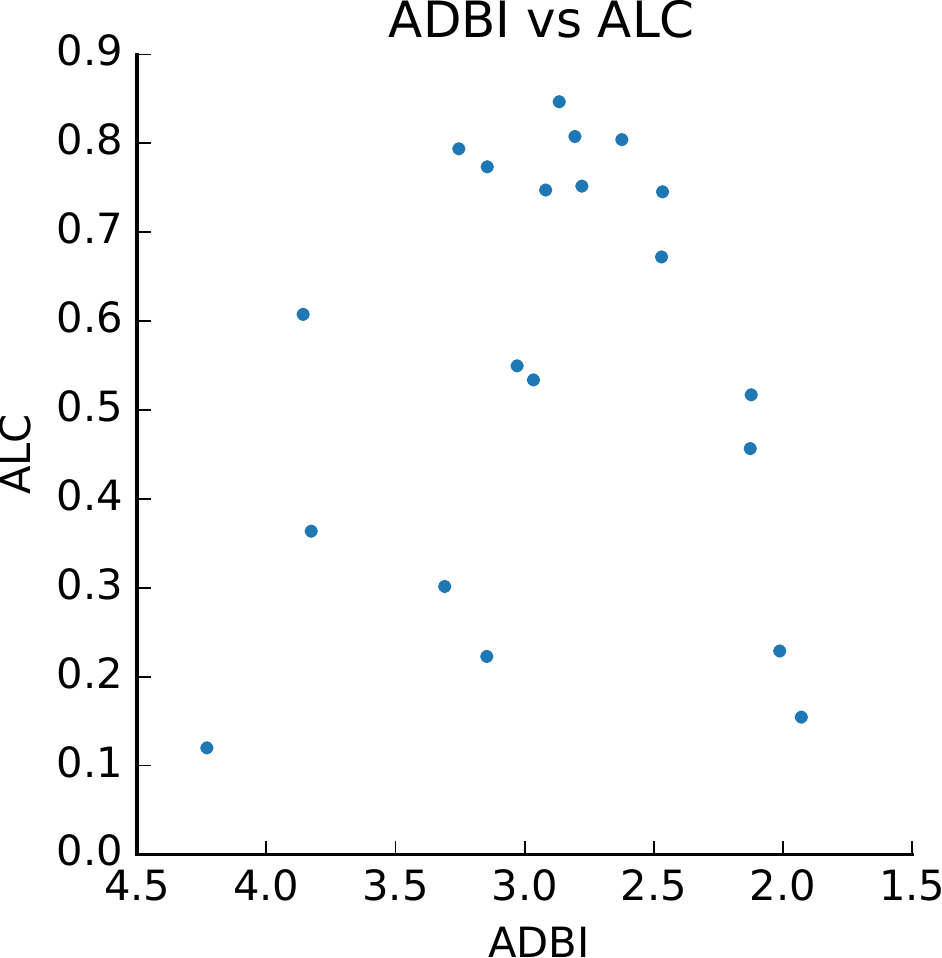}
    \caption{Correlation between ADBI and ALC. (The x-axis has been inverted.)}
    \label{fig:dbi_correlation}
\end{figure}

\begin{table}[t]
\begin{tabular}{llllll}
      & IMDB & WT & CC & S140 & avg \\ 
\hline
GloVe & 2.62 & 2.12 & 2.01 & 2.47 & \textbf{2.31} \\
fastText & 2.78 & 2.13 & 1.93 & 2.47 & 2.33 \\
BERT\textsubscript{cls} & 2.81 & 2.97 & 3.15 & 2.92 & 2.96 \\
BERT\textsubscript{all} & 2.87 & 3.03 & 3.31 & 3.25 & 3.11 \\
BoW & 3.14 & 3.83 & 4.23 & 3.86 & 3.76 \\
\hline 
\end{tabular}
\caption{ADBI for different representations and datasets (lower is better). Rows are sorted by average ADBI.}
\label{tab:dbi}
\end{table}
}

The right side of \tref{tab:results} shows the results for the same datasets and representations used for \textsc{Thas}.
GloVe induces the best clusters according to the ADBI metric. BERT\textsubscript{all} does not produce particularly good clusters despite being the strongest few-shot representation.
\fref{fig:correlations} (red crosses) and \tref{tab:correlation} show that there is a low correlation between the two metrics. This suggests that the geometric properties of the clusters alone can not explain few-shot performance.

\section{Related Work}
\label{related_work}


Representation choice has recently gained significant attention from the active learning (AL) community \citep{schroder_survey_2020, shnarch-etal-2022-cluster, zhang_active_2017}. Some work has attempted to quantify what representation is best when training the initial model for AL, which is usually referred to as the cold start problem \citep{lu_investigating_2019}. The importance of word embeddings has been also studied in the context of highly imbalanced data scenarios \citep{sahan_active_2021, naseem_comparative_2021, hashimoto_topic_2016, kholghi_benefits_2016}.
Most research conducted by the AL community on textual representations has focused on determining \emph{which} representations lead to higher performance for a given task. However, our paper aims to investigate \emph{why} a certain representation performs better in the few-shot scenario.

Our work, focused on examining properties of various textual representations, is closely related to recent research on evaluating the general capabilities of word embeddings. Many studies are interested in testing the behavior of such models using probing tasks that signal different linguistic skills \citep{conneau-etal-2018-cram, conneau-kiela-2018-senteval, marvin-linzen-2018-targeted, tenney_what_2019, miaschi_contextual_2020}. Others have targeted the capacity of word embeddings to transfer linguistic content \citep{ravishankar-etal-2019-probing, conneau-etal-2020-emerging}.

Looking at approaches that analyze the properties of representations directly, without intermediate probes, \citet{saphra-lopez-2019-understanding} developed a correlation method to compare representations during consecutive pre-training stages. Analyzing the geometric properties of contextual embeddings is also an active line of work \citep{reif_visualizing_2019, ethayarajh-2019-contextual, hewitt-manning-2019-structural}. While these previous works focus on analyzing representation properties independently, without considering a specific task, our study investigates the relationship between representations and task labels. We conduct a comparison between this relationship and the unsupervised analysis of representation properties.


Our work falls in line with broader research on the relationship between task and representation. 
\citet{yauney-mimno-2021-comparing} proposed a method to measure the alignment between documents and labels in a given representation space using a data complexity measure developed in the learning-theory community.
In the computer vision area, \citet{pmlr-v97-frosst19a} introduced a loss metric and investigated the entanglement of classes in the representation space during the learning process.
\citet{zhou-srikumar-2021-directprobe} proposed a heuristic to approximate the version space of classifiers using hierarchical clustering, highlighting how representations induce the separability of class labels, thereby simplifying the classification task. In contrast, our work specifically examines the few-shot performance and emphasizes the importance of unbalanced scenarios. We find that in these more realistic situations, the choice of representation plays a critical role, paving the way for advanced strategies in active learning.

\section{Conclusion}
\label{conclusion}

In this paper, we asked the question: What underlying property characterizes a good representation in a few-shot learning setting? We hypothesized that good representations are those in which the structure of the input space is well aligned with the label distribution.
We proposed a metric to measure such alignment: \textsc{Thas}. To test our hypothesis, we conducted experiments on several textual classification datasets, covering different classification tasks and label distributions (i.e. both balanced and unbalanced). We compared a range of word embedding representations as well as a baseline sparse representation.

Our results showed that when labeled data is scarce the best-performing representations are those where the input space is well aligned with the labels. Furthermore, we showed that the performance of a representation can not be explained by looking at classical measures of clustering quality.

The main insight provided in this work could be leveraged to design new strategies in active learning. The fact that good representations induce clusters of high purity at different granularities creates opportunities for wiser exploration of the representation space in an active manner. Similar to the work of \citet{dasgupta_hierarchical_2008}, we could employ the data dendrogram to guide this exploration.

\section*{Limitations}

In this paper, we focused on analyzing the properties of textual representations in the few-shot learning scenario. Its applicability to broader annotation scenarios could be presumed but is not supported by our empirical results. 

Our experimental setup is based on binary classification tasks using English datasets. While our approach is general and could be easily extended to multi-class scenarios, more work would be required to extend it to other more complex structured prediction settings such as sequence tagging. 

We see several ways in which this work could be extended. The most obvious extension consists of trying to generalize the notion of alignment to other tasks beyond sequence classification, such as sequence tagging. In this paper, we have used \textsc{Thas} to understand the quality of a given textual representation. However, since \textsc{Thas} is a function of a labeling and a representation, it could also be used to measure the quality of a labeling \citep{yan_cost_2018}, given a fixed representation. For example, this might be used in the context of hierarchical labeling, to measure which level of label granularity is better aligned with some input representation. 

The goal of this paper was to provide an explanation for the success of pre-trained word embeddings for text classification in the few-shot learning scenario. We believe that with our proposed methodology we have successfully achieved this goal. However, it should be clear to the reader that we do not provide a method for picking the best representation, i.e. for model selection. This is because our analysis requires access to labeled data and if labeled data is available the best way to select a model will be via cross-validation.

\section*{Acknowledgements}

This project has received funding from the European Research Council (ERC) under the European Union's Horizon 2020 research and innovation programme under grant agreement No 853459. The authors gratefully acknowledge the computer resources at ARTEMISA, funded by the European Union ERDF and Comunitat Valenciana as well as the technical support provided by the Instituto de Física Corpuscular, IFIC (CSIC-UV). This research is supported by a recognition 2021SGR-Cat (01266 LQMC) from AGAUR (Generalitat de Catalunya).

\bibliography{anthology,custom}
\bibliographystyle{acl_natbib}

\newpage
\appendix
\section{Appendix}

\subsection{\textsc{Thas} Implementation Details}
\label{thas_implementation}

The data dendrogram is obtained via hierarchical agglomerative clustering. More precisely, we use a bottom-up algorithm that starts with each sample as a singleton cluster and consecutively merges clusters according to a similarity metric and merge criterion until a single cluster is formed.

We apply \citeposs{ward_hierarchical_1963} method, which uses the squared Euclidean distance between samples and then minimizes the total within-cluster variance by finding consecutive pairs of clusters with a minimal increase. The clustering algorithm produces a list of merges that represent a dendrogram and can be traversed to generate a clustering partition for each value of $k$. It was implemented using Scikit-learn \citep{scikit-learn} and NumPy \citep{harris2020array}.

Expressed as a nearest-neighbor chain algorithm, Ward's method has a time complexity of $\mathcal{O}(n^2)$ \citep{10.1093/comjnl/26.4.354}.
\textsc{Thas} experiments have been performed using sub-samples of size $10$K and averaged over $5$ seeds.
Using $32$ CPUs and $16$GiB of RAM, each agglomerative clustering took on average $3.3$ minutes. Each task alignment curve took $3$ minutes on average. In contrast, DBI curves took $7.8$ hours on average.

\subsection{Experimental Details}
\label{experimental_details}

\paragraph{Datasets.}

\tref{tab:datasets} shows the statistics of the datasets used in this paper. They were extracted from HuggingFace Datasets \citep{lhoest-etal-2021-datasets}. For WikiToxic and CivilComments, we have applied a pre-processing consisting of removing all markup code and non-alpha-numeric characters.

\begin{table}[ht]
\centering
\begin{tabular}{lrrl}
\toprule
Dataset & Size & Prior & Task \\  
\midrule
IMDB & 50K & 50\% & sentiment \\  
WikiToxic & 224K & 9\% & toxicity \\  
Sentiment140 & 1.6M & 50\% & sentiment \\  
CivilComments & 2M & 8\% & toxic behav. \\  
\bottomrule
\end{tabular}
\caption{Datasets statistics with the number of samples, target (positive) class prior, and classification task.}
\label{tab:datasets}
\end{table}

\paragraph{Representations.}

The following is a detailed description of the text representations used in our experiments:

\begin{description}\setlength\itemsep{0em}
    \item BoW: this is a standard sparse term frequency bag-of-words representation.
    \item BERT\textsubscript{all}: word embeddings from \citeposs{devlin-etal-2019-bert} BERT\textsubscript{BASE} uncased model, average pooling of 2\textsuperscript{nd} to last layers and average pooling of all tokens.
    \item BERT\textsubscript{cls}: the same as above but using the \texttt{[CLS]} token alone.
    \item GloVe: \citeposs{pennington_glove_2014} word vectors pre-trained on Common Crawl with average pooling.
    \item fastText: word vectors from \citet{bojanowski_enriching_2017, joulin_bag_2016} pre-trained on Wikipedia with average pooling.
\end{description}

BERT representations were extracted using the HuggingFace Transformers library \citep{wolf-etal-2020-transformers} implemented in PyTorch \citep{paszke_pytorch_2019}.

\ignore{
BERT learning curves were averaged over $5$ sub-samples and initialization seeds and computed using $5$-fold cross-validation with the number of training epochs chosen over a maximum of $10$. In total, we performed $2400$ experiments ($4$ datasets, with and without pretraining, $5$ seeds, $10$ learning points, $5$ folds per cross validation plus $4 \times 2 \times 5 \times 10$ train and test learning points). The computation of each learning point took a total of $27$ minutes on average using a single Nvidia V100 GPU.
}

\paragraph{Models.}
The parameters for max-entropy learning curves were validated using $5$-fold cross-validation and the results averaged over sub-samples from $5$ seeds.

\subsection{Curves}
\label{curves}

\begin{figure*}[p]
    \centering

    \includegraphics[width=0.32\linewidth]{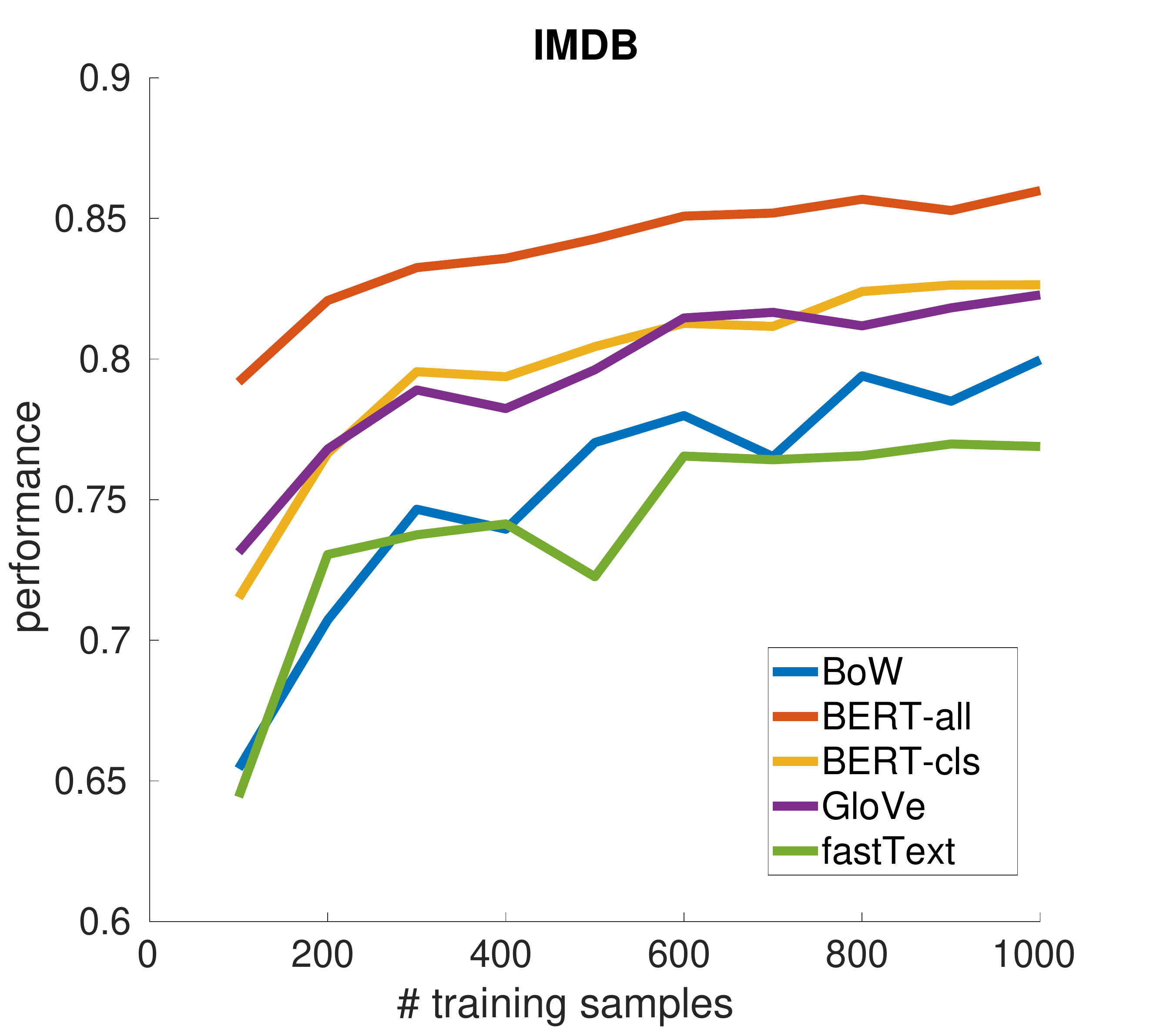}
    \includegraphics[width=0.345\linewidth]{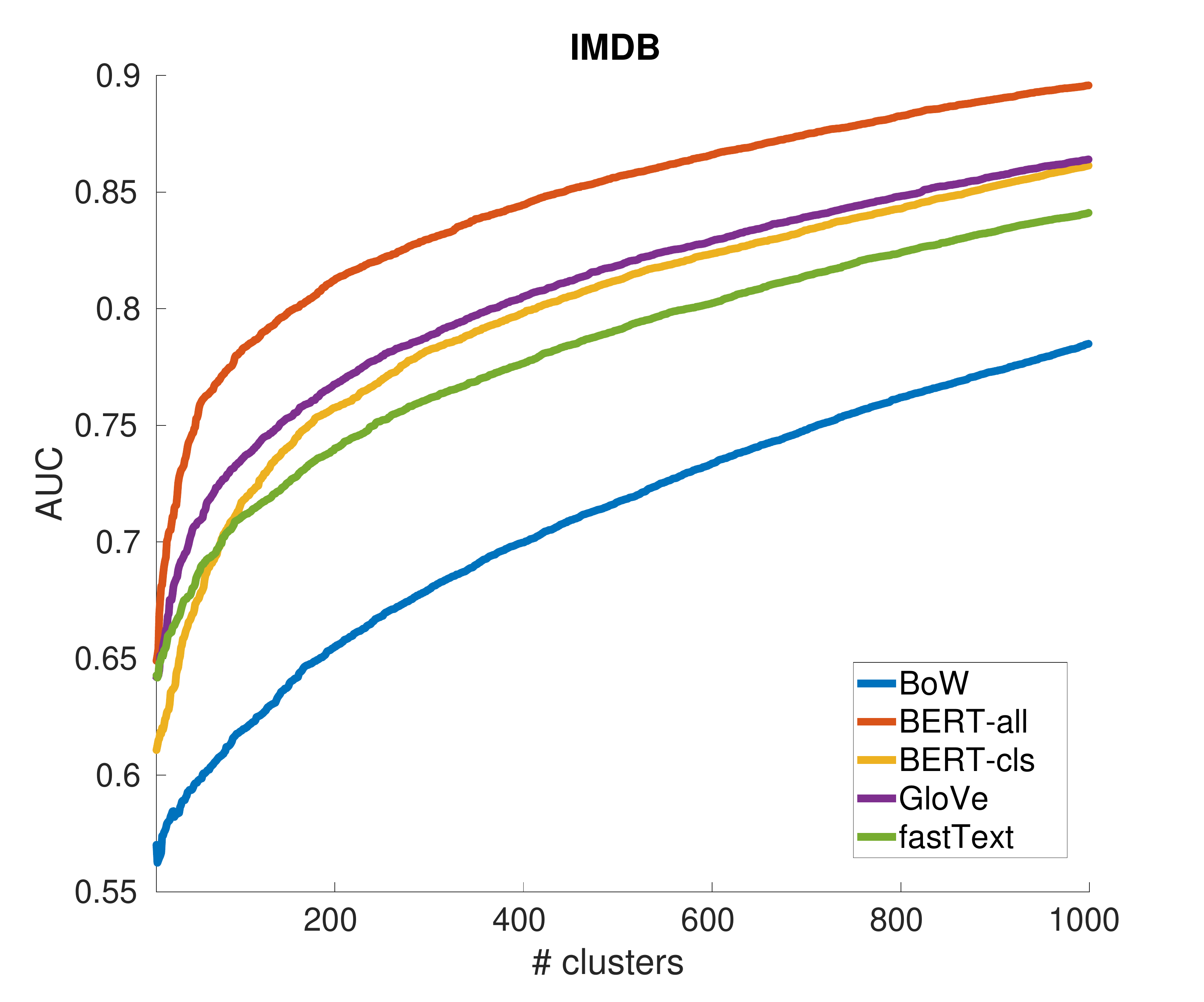}
    \includegraphics[width=0.32\linewidth]{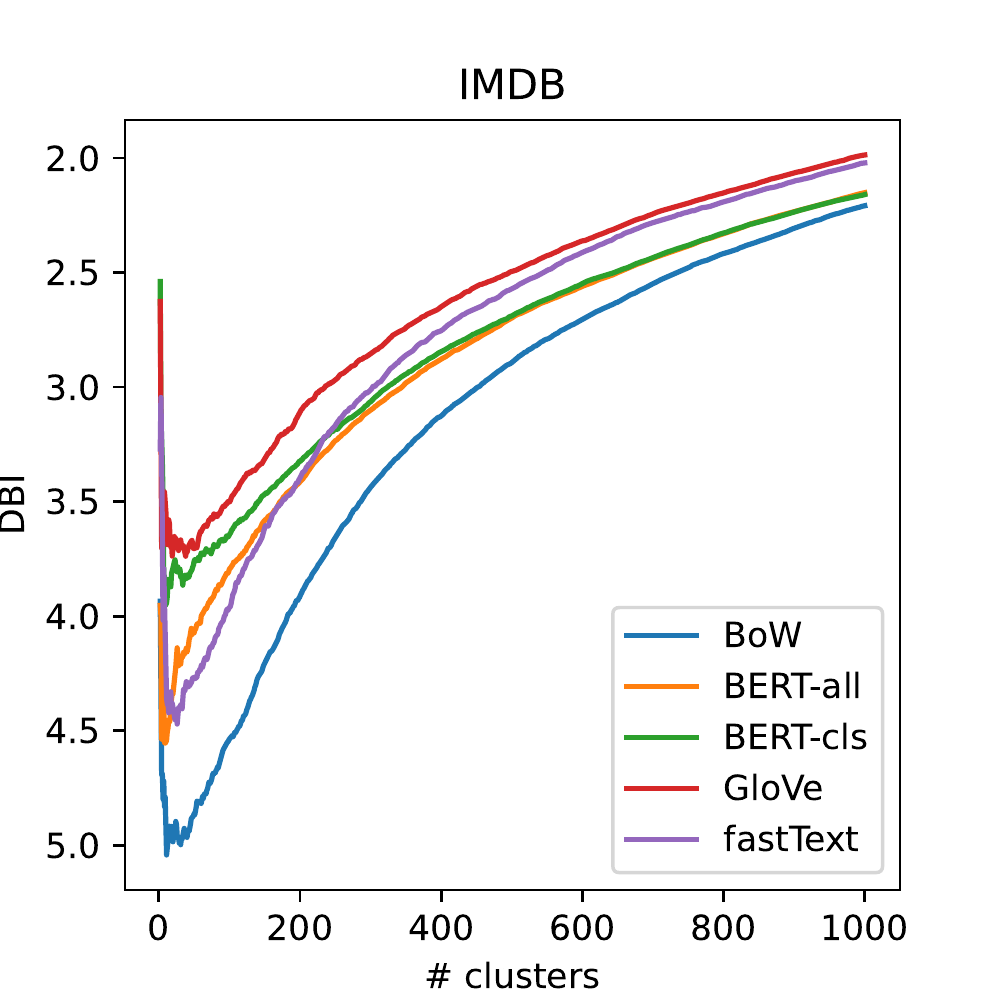}

    \includegraphics[width=0.32\linewidth]{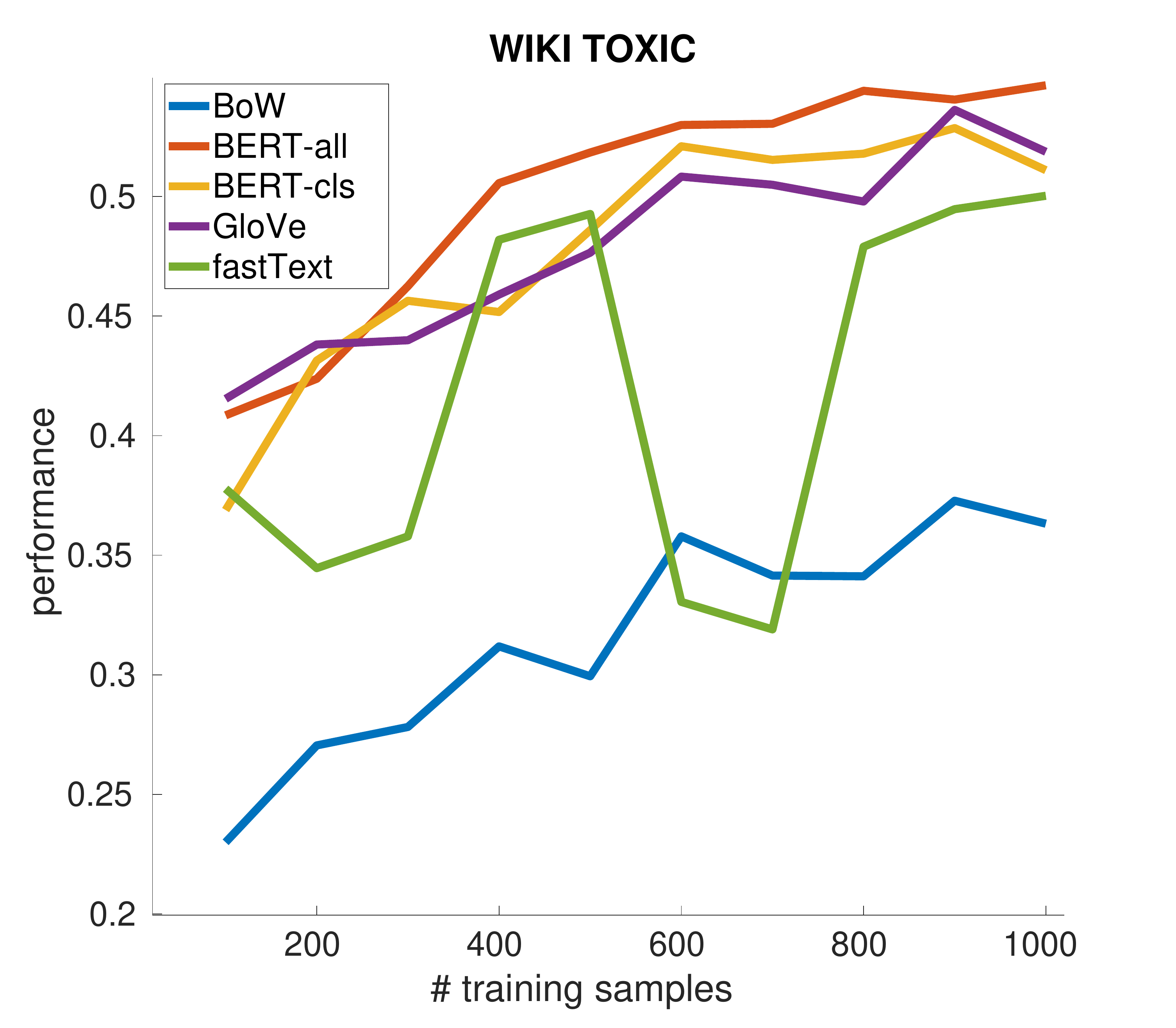}
    \includegraphics[width=0.345\linewidth]{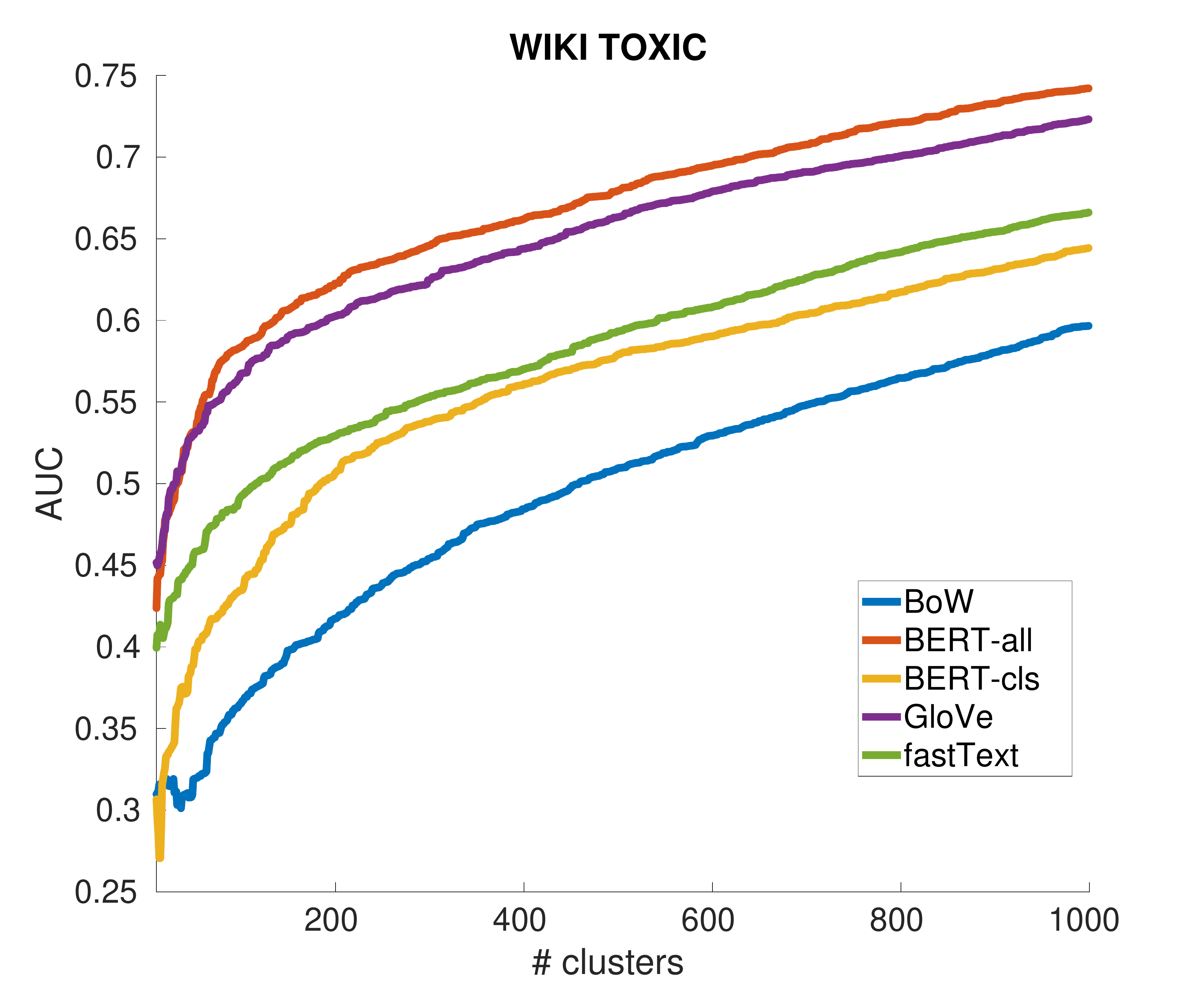}
    \includegraphics[width=0.32\linewidth]{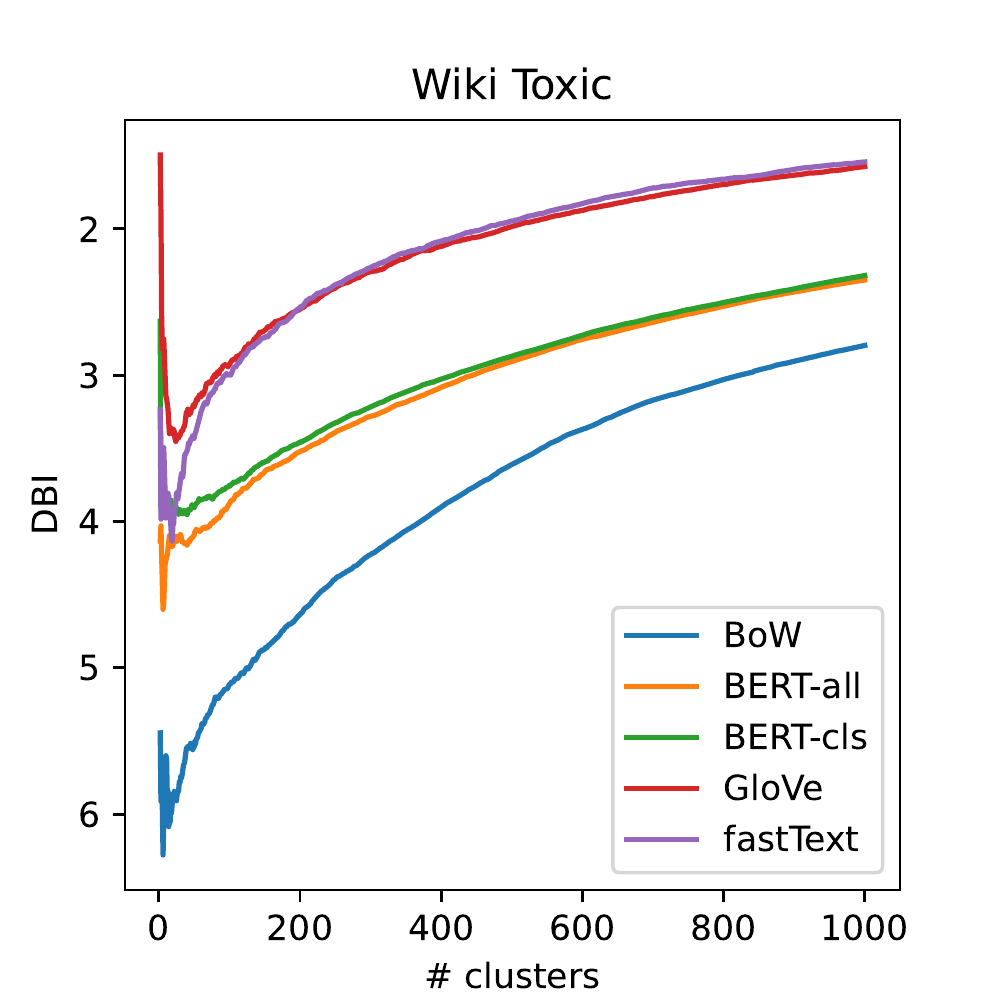}
    
    \includegraphics[width=0.33\linewidth]{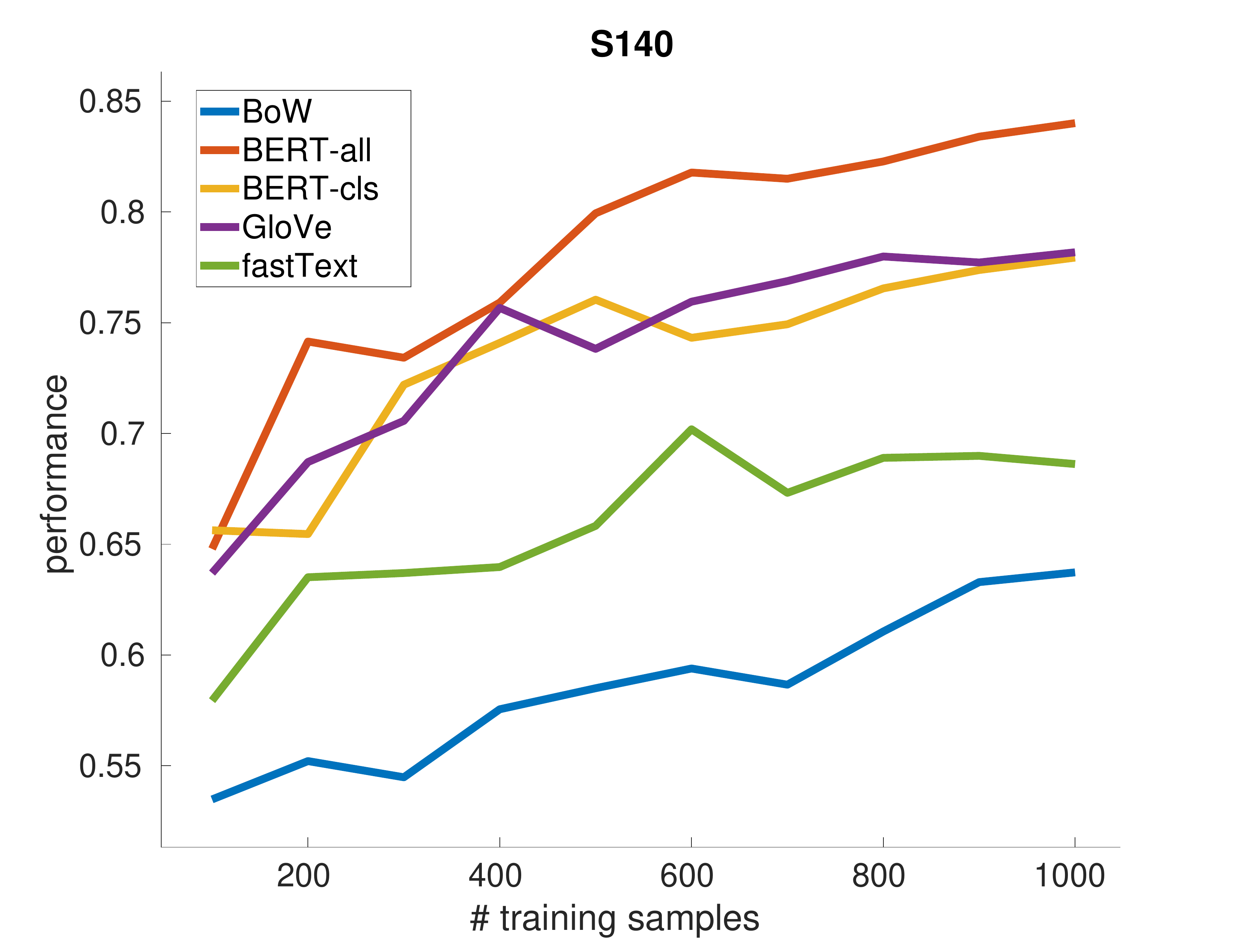}
    \includegraphics[width=0.33\linewidth]{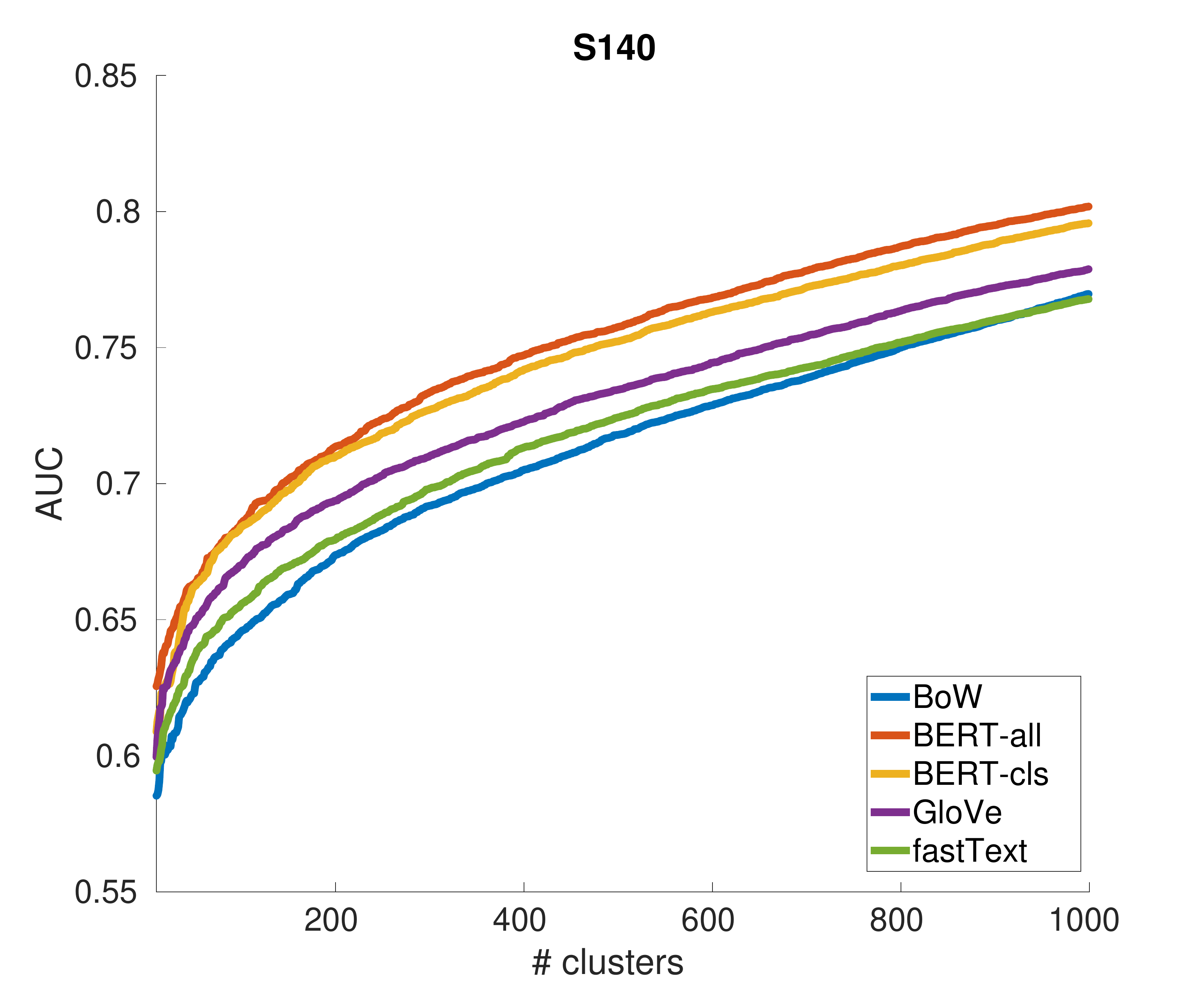}
    \includegraphics[width=0.32\linewidth]{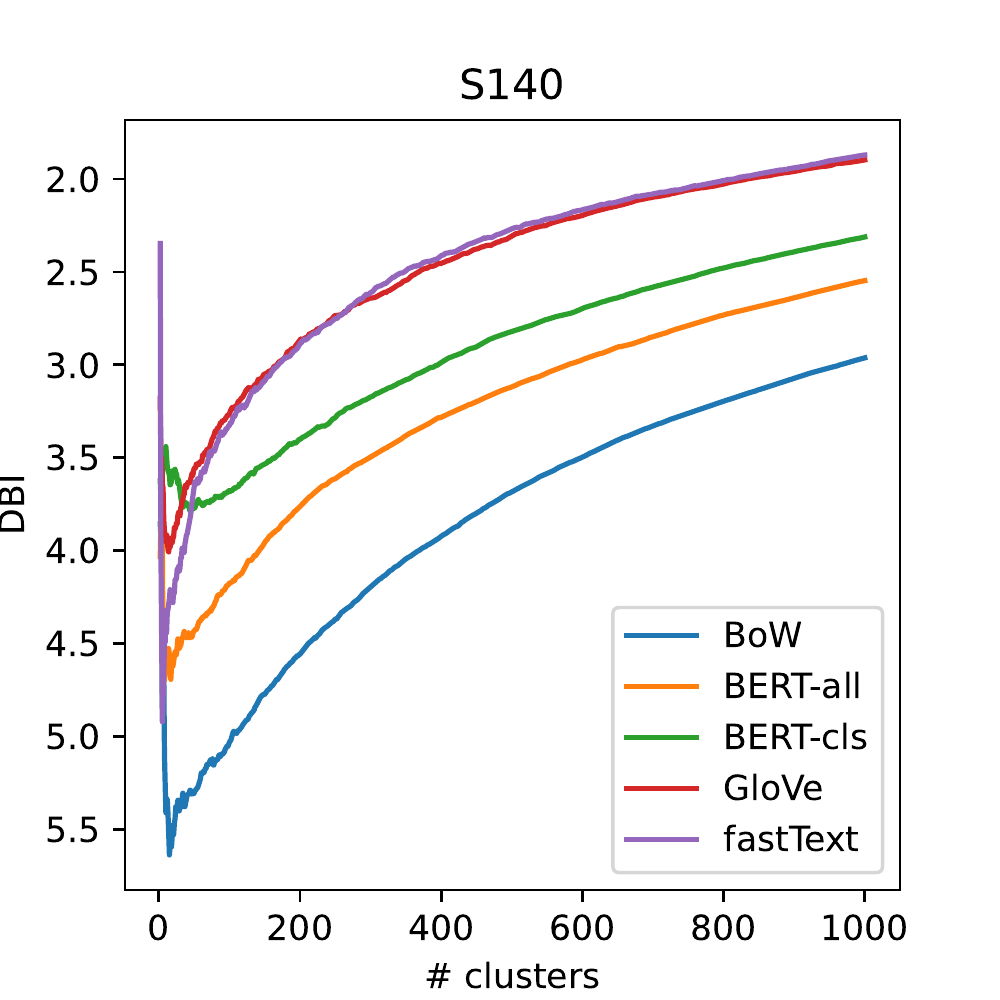}
    \vspace{1em}
    
    \includegraphics[width=0.33\linewidth]{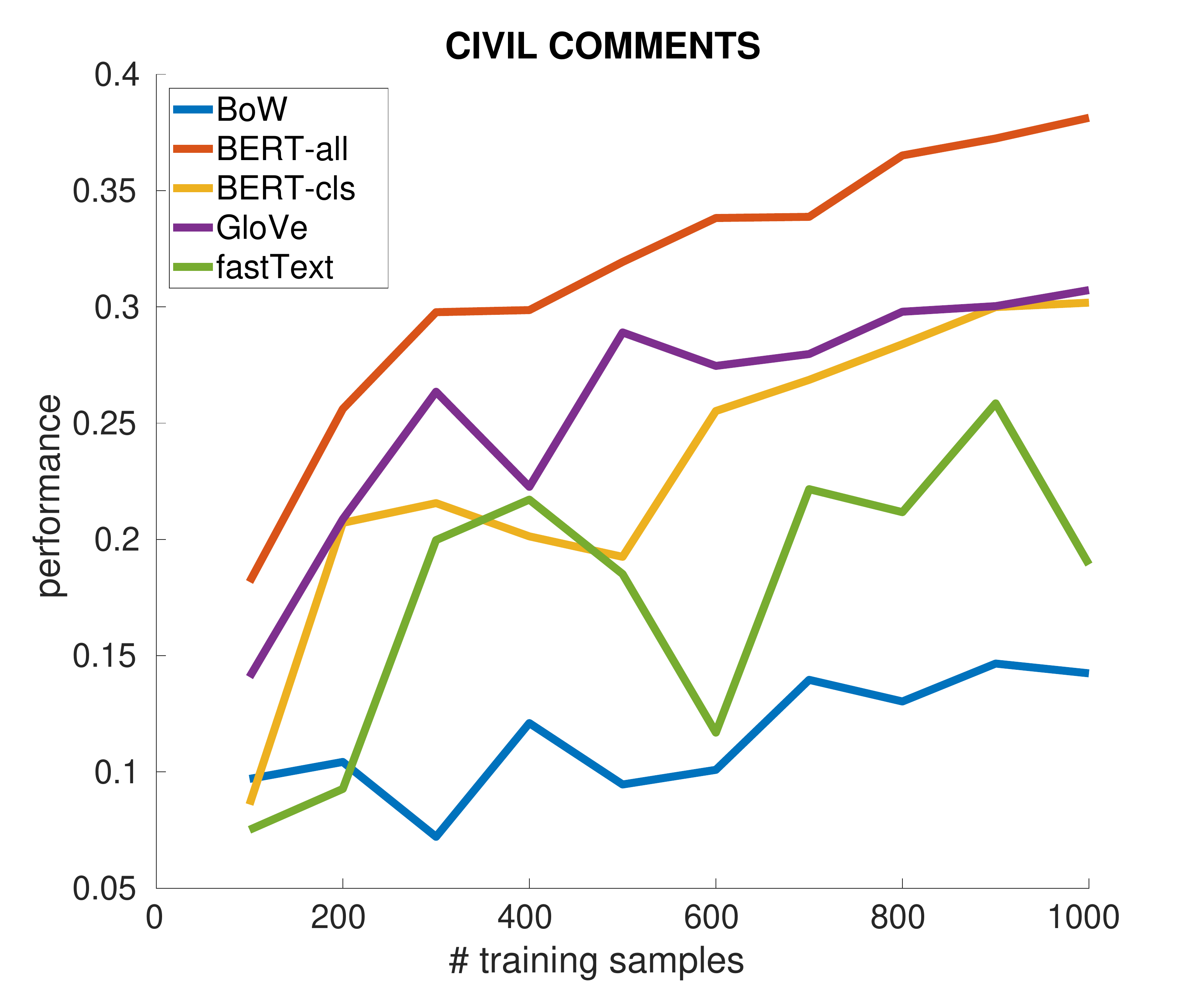}
    \includegraphics[width=0.33\linewidth]{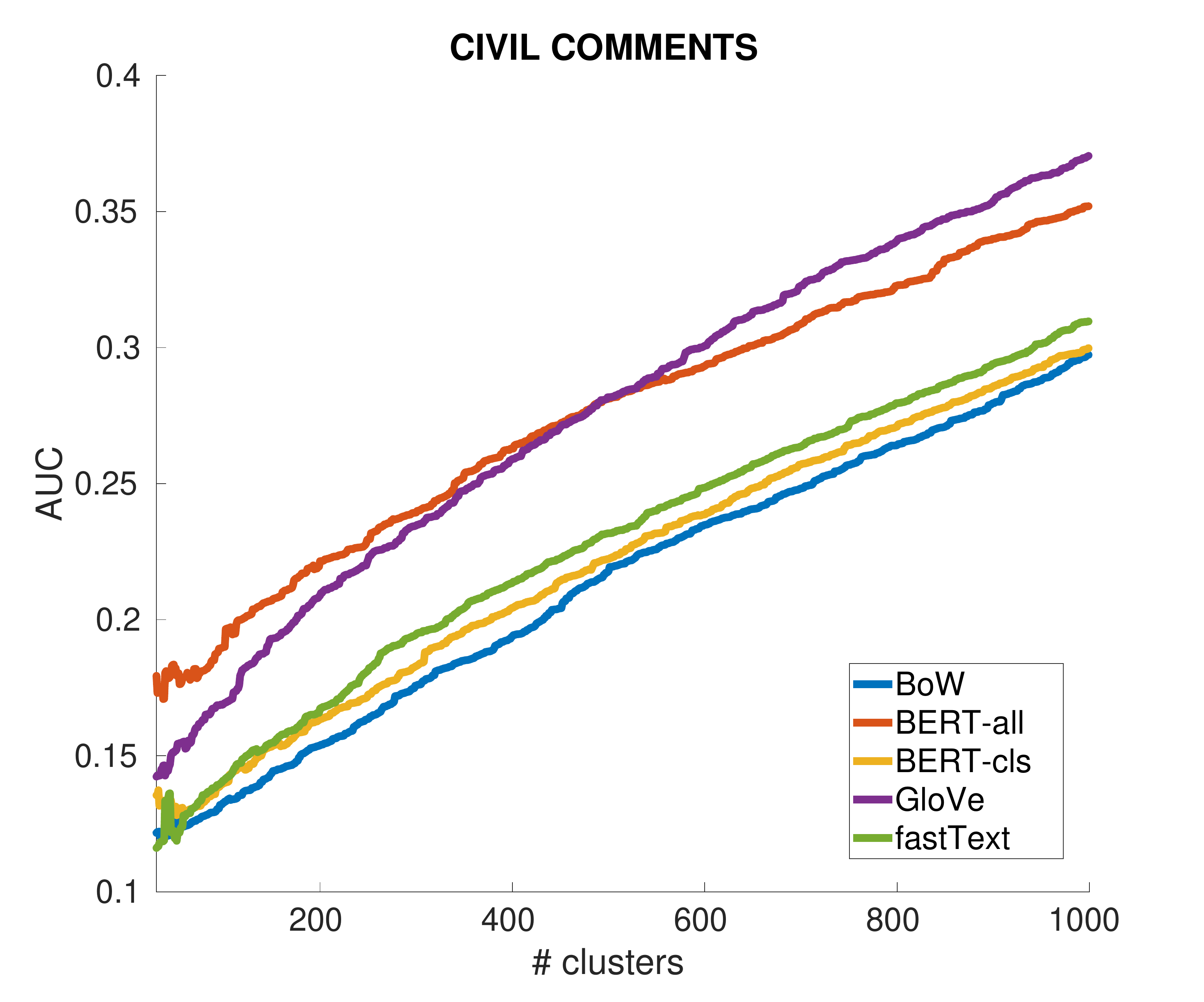}
    \includegraphics[width=0.32\linewidth]{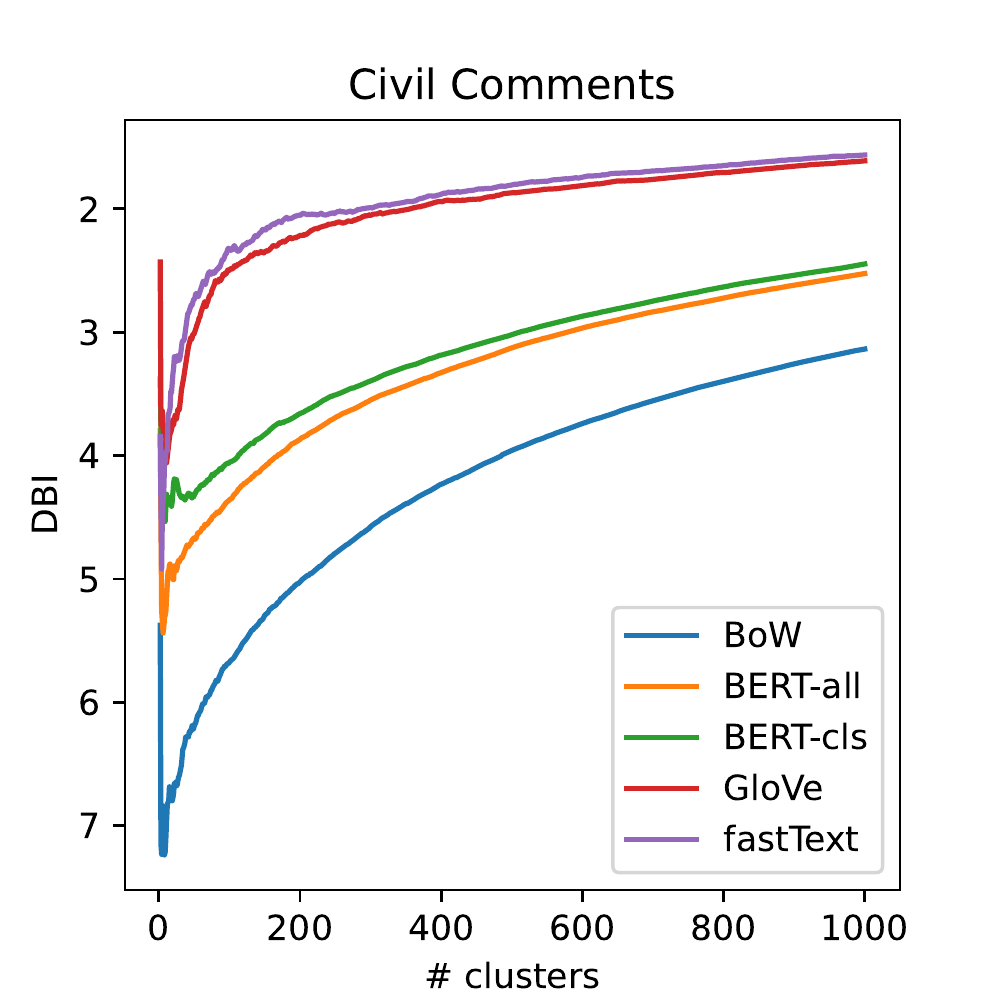}
    
    \caption{Learning curves (left), task hierarchical alignment curves (center), and DBI curves (right) for all the datasets: IMDB, WikiToxic, Sentiment140, and CivilComments.}
    \label{fig:curves}
\end{figure*}

\fref{fig:curves} presents the curves used to compute the main results in \sref{experiments}. The left column contains the learning curves used to compute the few-shot learning performance of the different datasets and representations.
The center column shows task alignment scores as a function of the number of clusters. \textsc{Thas} is computed as the area under these curves. The pre-trained word embeddings, in particular BERT, tend to achieve the best results. In the curves, they show higher values of alignment for a small number of clusters. The relative performance of the representations in the learning curves is paralleled in the task hierarchical alignment curves.
BERT\textsubscript{all} (i.e. using average pooling over all tokens) seems to be superior to BERT\textsubscript{cls} (i.e. using only the \texttt{[CLS]} token).

The right column in \fref{fig:curves} shows the DBI curves as a function of the number of clusters. These curves were used to compute the unsupervised clustering metric (ADBI) results presented in \sref{dbi}. As shown in the figure, these curves do not preserve the relative ranking we find in the corresponding learning curves.

\end{document}